\documentclass[letterpaper, 10 pt, conference]{ieeeconf}  
\IEEEoverridecommandlockouts                              
\usepackage{url}
\usepackage{hyperref}
\hypersetup{
    colorlinks=true,
    linkcolor=blue, 
    urlcolor=blue,  
    }
\usepackage{amsmath,amssymb}
\usepackage{graphicx,color}
\usepackage{booktabs}
\usepackage{multirow}
\usepackage{verbatim}
\usepackage{gensymb}
\let\labelindent\relax  
\usepackage{enumitem}
\usepackage{pifont}
\usepackage{xspace}
\usepackage{adjustbox}
\graphicspath{{figures/}}

\definecolor{britishracinggreen}{rgb}{0.23, 0.33, 0.19}

\newcommand{\action}{a}

\newcommand{\policy}{\pi}

\newcommand{\obs}{o}
\newcommand{\inx}{\mathbf{x}}
\newcommand{\demo}{\mathit{d}}

\usepackage[rightcaption]{sidecap} 
\usepackage{graphicx} 

\usepackage{array}
\makeatletter
\g@addto@macro{\endtabular}{\rowfont{}}
\makeatother
\newcommand{\rowfonttype}{}
\newcommand{\rowfont}[1]{
   \gdef\rowfonttype{#1}#1%
}
\newcolumntype{L}{>{\rowfonttype}l}
\newcolumntype{R}{>{\rowfonttype}r} 

\usepackage[font={footnotesize}]{caption} 

\title{\LARGE \bf
Towards More Generalizable One-shot Visual Imitation Learning 
}

\author{Zhao Mandi $^{*1}$, Fangchen Liu $^{*1}$, Kimin Lee $^{1}$, Pieter Abbeel $^{1}$ 
\thanks{$^{*}$Equal Contribution}%
\thanks{$^{1}$University of California, Berkeley, USA.}%
\thanks{ \url{https://sites.google.com/berkeley.edu/mosaic}}
\thanks{Correspondence to {\tt\small mandi.zhao@berkeley.edu}}%
}

\begin{document}

\maketitle
\thispagestyle{empty}
\pagestyle{empty}


\begin{abstract}
    A general-purpose robot should be able to master a wide range of tasks and quickly learn a novel one by leveraging past experiences.
    One-shot imitation learning (OSIL) approaches this goal by training an agent with (pairs of) expert demonstrations, such that at test time, it can directly execute a new task from just one demonstration.
    However, so far this framework has been limited to training on many variations of one task, and testing on other unseen but similar variations of the same task. 
    In this work, we push for a higher level of generalization ability by investigating a more ambitious multi-task setup. We introduce a diverse suite of vision-based robot manipulation tasks, 
    consisting of 7 tasks, a total of 61 variations, and a continuum of instances within each variation.
    For consistency and comparison purposes, we first train and evaluate single-task agents (as done in prior few-shot imitation work). We then study the multi-task setting, where multi-task training is followed by (i) one-shot imitation on variations within the training tasks, (ii) one-shot imitation on new tasks, and (iii) fine-tuning on new tasks. Prior state-of-the-art, while performing well within some single tasks, struggles in these harder multi-task settings. To address these limitations,
    we propose MOSAIC (\textbf{M}ulti-task \textbf{O}ne-\textbf{S}hot \textbf{I}mitation with self-\textbf{A}ttention and \textbf{C}ontrastive learning), which integrates a self-attention model architecture and a temporal contrastive module to enable better task disambiguation and more robust representation learning. Our experiments show that MOSAIC outperforms prior state of the art in learning efficiency, final performance, and learns a multi-task policy with promising generalization ability via fine-tuning on novel tasks. 
\end{abstract}

\section{INTRODUCTION}

\begin{figure*}[ht!]
\center
\includegraphics[width=0.9\textwidth]{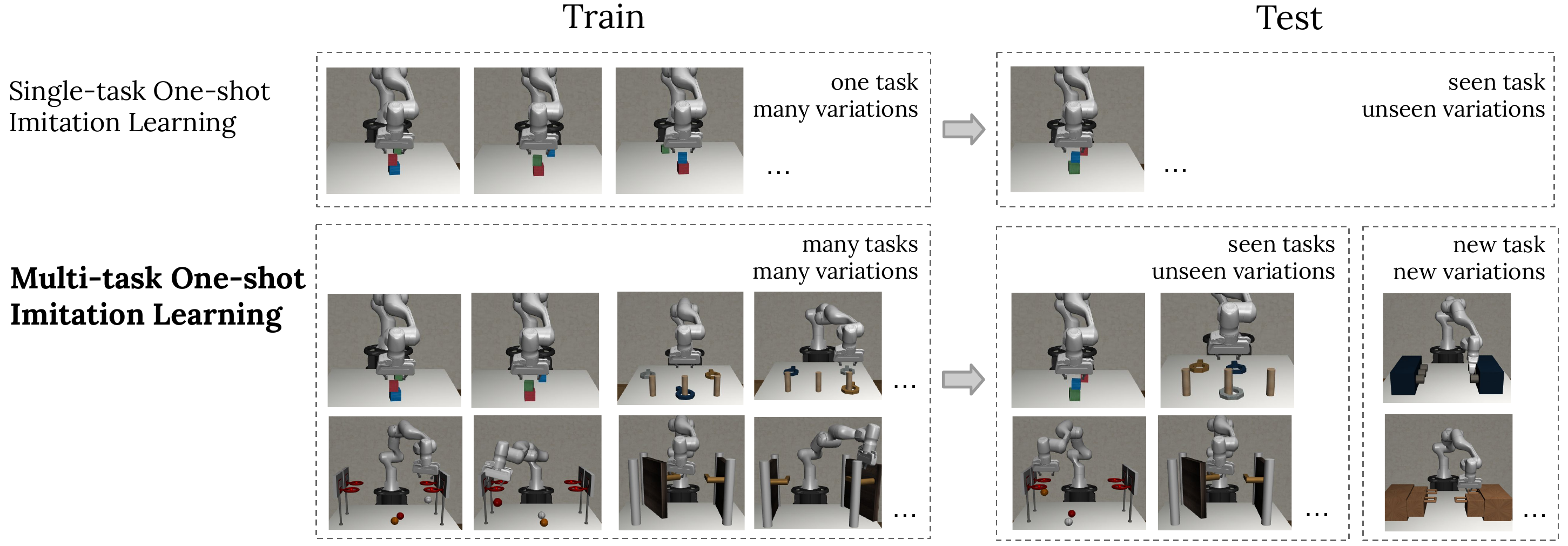}
\caption{ 
Illustration of train/test split in prior and our proposed settings. In contrast to prior work in one-shot imitation learning which add more variations to a single task, we propose to train on multiple distinct tasks along with all of their variations, and test not only on multiple trained tasks, but also on novel tasks that were never seen during training.
} 
\vspace*{-6pt}
\label{fig:fig1}
\end{figure*}
 
Humans can learn to complete many tasks and quickly adapt to a new situation based on past experiences. We believe robots should also be able to learn a variety of tasks and acquire generalizable knowledge, which can then be transferred to quickly and efficiently learn a novel task.

One-shot imitation learning (OSIL) is a popular training framework for this purpose: an agent is trained to perform multiple tasks, each is described by an expert demonstration to provide context. First proposed in \cite{duan2017one}, the framework has been extended to different tasks and visual inputs \cite{dasari2020transformers,finn2017one,james2018task,yu2018one}. However, these prior works tend to assume a very strong similarity between train and test. For example, a typical setting is where at training time the agent learns to build a block configuration that matches the block configuration in a preceding demo, and then at test time the agent is again requested to stack blocks, with variation just stemming from which block is in which position of the stack and the starting locations of blocks on the tabletop. Another typical setting is moving objects from a table top to a set of bins, where at test time the task will again be moving the same objects from tabletop to bins, with variation just stemming from which object goes to which bin and starting locations of the objects.

In this work, we propose to expand this narrow, single-task setting with a more significant distinction between train and test. Concretely, we build 7 robot manipulation environments: Door, Drawer, Press Button, Stack Block, Basketball, Nut Assembly, Pick \& Place, based on simulation framework from Robosuite v1.1~\cite{zhu2020robosuite} and MetaWorld~\cite{yu2019meta}. Terminology-wise, we will consistently refer to them as tasks. Within each task, there are ``variations'', which capture differences in which block goes on top of which block, or which object goes into which bin, etc (see Figure \ref{fig:fig3} for visualizations of all 61 variations). Within one fixed variation of a task, there is also a continuum of instances, corresponding to all possible initial states of the various objects. We illustrate this setup with the row ``Multi-task One-shot Imitation Learning" in Figure \ref{fig:fig1}. 



We evaluate representational and generalization capability through three settings: (i) one-shot imitation on variations within the multi-task training regime, (ii) one-shot imitation on new tasks; and (iii) fine-tuning on new tasks. As a first step in our investigation, we study the performance of prior state-of-the-art methods \cite{dasari2020transformers,yu2018one}. We observe that, while performing well in the prior single-task settings, these methods largely fail to handle our proposed multi-task setup. This suggests a great opportunity for novel research towards improving the generalization ability of few-shot imitation methods.
 
In addition to identifying this challenge for more generalizable OSIL, we also propose a new approach that shows significant performance gain over prior state of the art.
Concretely, we investigate the hypothesis that prior methods fall short in the multi-task settings due to (i) poor representations that do not generalize well to new tasks; (ii) a lack of proper inductive bias in the model architecture, which prevents accessing the one demonstration from a new task.  To address these challenges, we introduce \textbf{MOSAIC}: \textbf{M}ulti-task \textbf{O}ne-\textbf{S}hot imitation with self-\textbf{A}ttent\textbf{I}on and \textbf{C}ontrastive learning, which incorporates two key components:  (i) a new temporal contrastive loss objective to provide additional supervision for representation learning; (ii) a self-attention policy model architecture for extracting contextual information in the demonstration. Experimental results show significantly improved performance of our method over the prior state of the art. 

Key contributions of this paper can be summarized as the following:
\begin{itemize} 
    \item We introduce a simulated robotic manipulation benchmark that spans 7 tasks and a total of 61 task variations. Its codebase is publicly released to facilitate future research.
    \item We propose a more challenging setup for one-shot imitation learning: 1) train an agent on multiple distinct tasks and test on the seen task variations; 2) train on multiple distinct tasks and test on completely new tasks, via direct one-shot execution or fine-tuning. We investigate prior state-of-the-art methods under these conditions and observe clear room for improvement.
    \item We propose our method MOSAIC, which combines a self-attention model architecture and a temporal contrastive objective. We experimentally demonstrate its superior performance over baselines, and show its promising ability at being fine-tuned to learn a new task efficiently. 
\end{itemize}

\section{RELATED WORK}
{\bf Imitation learning}. There are two main approaches for imitation learning (IL): inverse reinforcement learning (IRL)~\cite{abbeel2004irl,ng2000irl,ho2016generative} which finds a cost function under which the expert is uniquely optimal, and behavioral cloning (BC)~\cite{bain1995framework,pomerleau1991efficient} that predicts expert actions from state observations as a supervised learning problem. 
Recent advances in IL have enabled agents to perform various robotic control tasks, such as locomotion~\cite{peng2020learning, ho2016generative}, self-driving~\cite{bojarski2016end, pomerleau1998autonomous}, video games~\cite{aytar2018playing,ross2011reduction}, and manipulation~\cite{pook1993recognizing,sweeney2007model,young2020visual}. However, a majority of these applications assumes a close match between train and test environment. This has the disadvantage of learning without the ability to transfer knowledge to new situations, and lacks the opportunity for a human to instruct the agent with a new task at test time. 
 
{\bf One-shot imitation learning}. To address these limitations, one-shot imitation learning (OSIL), first proposed in \cite{duan2017one}, trains an agent to intake both one successful demonstration and the current observation, and predict the expert's action. Later work extended OSIL to observe visual inputs: \cite{finn2017one} applies the Model-Agnostic Meta-Learning algorithm (MAML)~\cite{finn2017model} to adapt policy model parameters for new tasks; TecNets \cite{james2018taskembedded} applies a hinge rank loss to learn explicit task embeddings; DAML \cite{yu2018one} adds a domain-adaptation objective to MAML to use human demonstration videos; \cite{dasari2020transformers} improves policy network with Transformer architecture~\cite{vaswani2017attention}. Another line of work learns modular task structures that can be reused at test time \cite{xu2018neural} \cite{huang2019neural} \cite{Huang2019ContinuousRO}, but outputs of these symbolic policies are highly abstracted into semantic action concepts (e.g. ``pick", ``release") that assume extensive domain knowledge and human-designed priors. 

However, prior OSIL work has been limited to a {\em single-task} setup and mainly tests a model on a slightly different instance (e.g. different object pose) of the previously-seen task variations. For example, \cite{finn2017one} and \cite{james2018taskembedded} experimented with 3 separate settings: simulated planer reaching (with different target object colors), simulated planer pushing (with varying target object locations), and real-robot, object-in-hand placing (onto different target containers). In contrast, we consider a more difficult multi-task setup, where agent needs to perform well across more diverse and distinct tasks, and generalize not only to new instances of all the seen variations, but also to completely novel tasks.

{\bf Multi-task Imitation Learning for Robotic Manipulation}
Our work falls under the broader category of imitation learning multiple robot manipulation tasks \cite{zhou2020watch}\cite{lynch2019play}\cite{singh2020scalable}. The term ``multi-task" has varying definitions across this space of literature. Some work define stacking different block combinations as different tasks, whereas we define them as variations of the same task. Tasks that are sufficiently distinct, such as object pushing versus grasping, are sometimes called ``task families" \cite{zhou2020watch}, where a ``multi-task" policy is trained with only one family, and novel object configurations are named ``new tasks" to test generalization. Recent work \cite{lynch2021language} \cite{NEURIPS2020_9909794d} also explored language conditioning for different interact-able objects as tasks, where act-once word embeddings are used for disambiguation, and shows generalization ability at sequentially executing trained tasks to achieve longer-horizon test-time tasks. Concurrently, BC-0 \cite{jang2021bc0} reports 100 distinct manipulation tasks for zero-shot imitation learning, where the 100 ``tasks" fall into only 9 underlying skills and 6-15 different objects, and the ``unseen" tasks are object arrangements that are excluded from training. 


{\bf Unsupervised/self-supervised representation learning}. 
Recently, several unsupervised/self-supervised representation learning methods have been proposed to improve the performance in learning from visual inputs~\cite{aytar2018playing,laskin2020curl,stooke2020decoupling,schwarzer2021dataefficient}. 
\cite{aytar2018playing} solves hard exploration environments like Atari’s Montezuma's Revenge~\cite{bellemare2013arcade} by using self-supervised representation to overcome domain gaps between the demonstrations and an agent's observation.
CURL~\cite{laskin2020curl} and ATC~\cite{stooke2020decoupling} showed that sample-efficiency can be significantly improved by applying contrastive learning~\cite{chen2020simple,henaff2020data, he2019momentum} to reinforcement learning. 
In this paper, we show that contrastive learning also provides large gains in one-shot imitation learning.


\section{PROBLEM SETUP}
We extend the framework of one-shot imitation learning (OSIL)~\cite{duan2017one} to a challenging multi-task setup. We categorize a set of semantically similar variants of a single task as ``variations": for example, for each variation of the Pick \& Place task, the agent should pick up one of 4 differently shaped objects, and place it in one of the 4 bins, resulting in 16 variations in total. Following this definition, prior work~\cite{duan2017one,dasari2020transformers, finn2017one, james2018task,yu2018one} on one-shot imitation learning evaluate agents with a single task, as illustrated in Figure~\ref{fig:fig1}. 
 
Consider $n$ different tasks, $\{\mathcal{T}_1, \mathcal{T}_2, \dots, \mathcal{T}_n$\}, where each task $\mathcal{T}_i$ contains a set of variations $\mathcal{M}_i$. For each task, the training dataset $\mathcal{D}_i$ contains paired expert demonstrations and trajectories from multiple variations: $\mathcal{D}_i = \{ (\demo_m, \tau_m), m \in \mathcal{M}_i \} $. The demonstrator provides a video $\demo_m = \{ \obs_0, \cdots, \obs_{T} \}$, and the policy is trained to intake $\demo_m$ and imitate an expert trajectory $\tau_m = \{ \left(\obs_0, \action_0 \right), \cdots, \left(\obs_T, \action_T \right) \}$. While expert trajectories require both actions and observations, only video inputs are required as demonstrations, thus the demonstrator can be any other robot or even human. For all our experiments, the demonstrator robot has a different arm configuration (i.e. Sawyer Arm) than the imitator agent (i.e. Panda Arm).
 
Given training datasets $\mathcal{D} = \{ \mathcal{D}_1, \cdots, \mathcal{D}_n\}$ of those $n$ tasks, we optimize a demonstration-conditioned policy $\policy_\theta (\action_t|\obs_t, \demo)$, parameterized by $\theta$, that takes an expert video and the current observation as input and takes an action at each time-step $t$.  At test time, the model is provided one demonstration $\demo_{\text{test}}$ of one variation of a task $m_{\text{test}}$, paired with observations $m_{\text{test, t}}$. Note that $m_{\text{test}}$ can be an unseen variation of either one of the trained tasks, or a never-seen task excluded from training.
\section{MAIN METHOD}
\begin{figure*}[t]
\center
\includegraphics[width=0.9\textwidth]{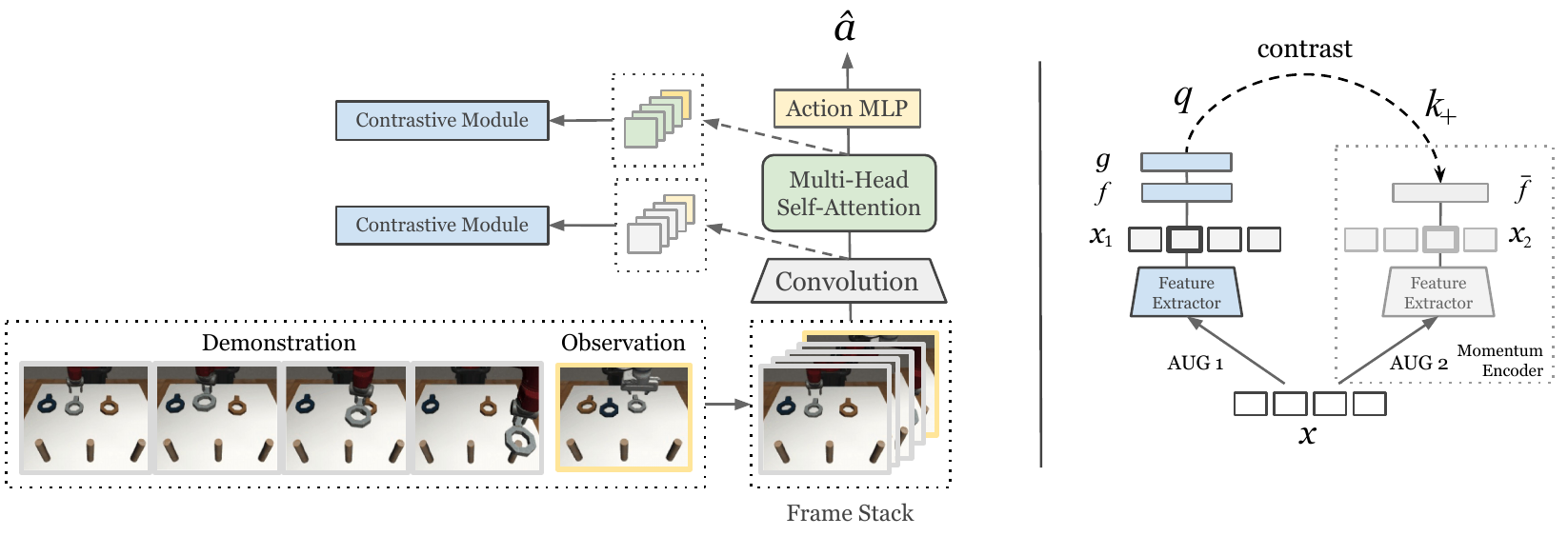}
\caption{
\small
Illustration of our overall network architecture (left) and contrastive module (right). The policy network takes in a stack of demonstration video frames and state observation images, and predicts the expert action at each time-step. A temporally-contrastive loss is applied in auxiliary with the policy's behavior cloning loss. Note that contrastive loss can be applied to features from either the convolution module or self-attention module. The model in gray box is gradient-free, and only receives parameter updates from its online counterpart. 
} 
\vspace*{-6pt}
\label{fig:fig2}
\end{figure*} 

In this section, we describe in details our approach to policy model architecture (Section~\ref{sec:arch}), self-supervised representation learning via a contrastive module (Section~\ref{sec:ssl}) and the action imitation loss objective (Section~\ref{sec:act}).

\subsection{Network Architecture} \label{sec:arch}
 
Figure~\ref{fig:fig2} provides an overview of our model pipeline: a CNN backbone is followed by a multi-head self-attention module~\cite{vaswani2017attention} and an MLP to make action predictions. 

{\bf Visual features} Given a batch of $B$ inputs containing $T_d$ expert video frames and $T_o$ agent observations, a CNN backbone encodes each frame into $C$ channels of size $H\times W$ feature maps, resulting in demonstration features $\inx_{\demo}$ with size $[B, C, T_d, H, W]$, and observation features $\inx_{o}$ with size $[B, C, T_o, H, W]$.
To preserve spatial and temporal information, both features are flattened along the last 3 dimensions and added with sinusoidal encodings \cite{vaswani2017attention}, then re-shaped into the original size.

{\bf Self-attention module} We use multiple self-attention layers that model the underlying relationship between the sequence of representations $\inx_{\demo}$ and $\inx_{o}$. We adopt the non-local self-attention block in \cite{wang2018non}, and make it to a multi-head version as \cite{dasari2020transformers}. Specifically, key, query and value tensors are generated from three separate 3D convolution layers, which are then flattened along the time and space dimensions to compute spatio-temporal attention by each head individually.
Formally, given temperature parameter $\tau$, key $K_j$, query $Q_j$ and value $V_j$, the attention head $j$, the attention operation is computed as:
\begin{align*}
    A_j = \mathbf{softmax}(Q_j K_j^\top / \tau)V_j.
\end{align*}
Outputs $A_j$ from each head are concatenated and projected to the original feature size by another 3D convolution, as used in \cite{wang2018non}. $\inx_{\demo}$ will first pass through a self-attention module to get $\inx_{\demo}^{\text{attn}}$. Then, every frame in $\inx_{o}$ will compute self-attention with both $\inx_{\demo}^{\text{attn}}$ and itself, which in effect calculates: (1) the spatial self-attention in each observation frame and (2) spatio-temporal cross-attention on the demonstration. The resulting $\inx_{o}^{\text{attn}}$ will be used to predict action. 

\subsection{Contrastive Representation Learning}
\label{sec:ssl}


Our method bases off the intuition that, representations for two nearby frames from the same video clip should be similar, whereas frames from different tasks or variations should be drawn apart.
For each frame in a video, we maximize its feature similarity with a randomly selected, temporally close-by frame. 
Specifically, we take an input batch, and obtain its two ``view''s by two separate data augmentations.
The model encodes the first view into $\inx_1$, and a target model encodes the second to get $\inx_2$, which is gradient-free and receives parameter updates solely from its online counterpart.  
Lastly, $\inx_1$ and $\inx_2$ are separately passed through a linear projector $f$: $z_1 = f(\inx_1)$, and its target $\bar f$: $z_2 = {\bar f}(\inx_2)$. For every feature frame in $z_1$, we select a nearby feature frame in $z_2$ as positive. We then maximize the similarity between each anchor $q = g(z_1)$, and its positive $k_+ = z_2$, via the InfoNCE loss from \cite{oord2019representation}, where
$g$ is another linear projector, also named as {\em predictor} in prior work \cite{grill2020bootstrap}. 

We follow \cite{oord2019representation, Hnaff2020DataEfficientIR} to model embedding similarity as bilinear product, calculated with a projection matrix $W$.  
Formally, with total frame count $F = B(T_d + T_o)$, treating every other $k$ in the batch as negatives, the contrastive loss at each $q$ is expressed as:
 
\begin{align}
    \mathcal{L}_{\text{Rep}} = \log \frac{\exp \left( q^T W k_+ \right)}
    { \exp \left(q^T W k_+ \right) + \sum_{i=1}^{F-1} \left( q^T W k_i \right) }
\end{align}

One may view the convolution backbone and self-attention layers as one combined feature extractor, therefore the above contrastive loss can be applied to either before or after the self-attention layers as shown in Figure~\ref{fig:fig2}. Moreover, our contrastive module differs from \cite{laskin2020curl} in the new temporal contrast strategy: for frame feature $x_t$ at timestep $t$, it contrasts with a random nearby frame selected from $x_{t-k}$ to $x_{t+k}$ in its augmented counterpart, whereas prior work \cite{stooke2020decoupling} uses a fixed-step future frame. We provide ablation experiments in Appendix to provide additional insights on details of our contrastive objetive implementation. 
 
\subsection{Policy Learning} \label{sec:act}
Our objective is to learn a policy $\policy_\theta (\action_t|\obs_t, \demo)$ which takes current image observation and a demonstration video as inputs, and predicts the action distribution to successfully finish the task. 

To enable learning a potentially multi-modal policy that excels across many tasks, we adopt the same solution used in \cite{dasari2020transformers, lynch2020learning, salimans2017pixelcnn++}, which discretizes the action space into 256 independent bins along every dimension, and parameterize the policy using a mixture of discretized logistic distribution. As described in Section~\ref{sec:arch}, the self-attended observation features $\inx_{o}^{\text{attn}}$ will pass through the action MLP, to predict the mean $\mu_i$, scale $s_i$ and mixing weight $\alpha_i$ for each discretized logistic distribution. The behavior cloning training loss is the negative log-likelihood:
\begin{align}
    \mathcal{L}_{\text{BC}} = - \log \left(\sum_{i=1}^m \alpha_i(\inx_{o}^{\text{attn}}) P\left(a_t, \mu_i(\inx_{o}^{\text{attn}}), s_i(\inx_{o}^{\text{attn}})\right)\right) 
    \label{loss:bc}
\end{align}
Where $P\big(a_t, \mu_i, s_i\big) = \sigma(\frac{a_t+0.5-\mu_i}{s_i}) - \sigma(\frac{a_t-0.5-\mu_i}{s_i})$, $\sigma$ is the logistic sigmoid function. 
At the inference time, given $o_t$, the action is sampled from the predicted distribution:
\begin{align}
    a_t \sim \sum_{i=1}^m \alpha_i(\inx_{o_t}^{\text{attn}}) \text{ logistic}\left(\mu_i(\inx_{o_t}^{\text{attn}}), s_i(\inx_{o_t}^{\text{attn}})\right) 
\end{align}
In addition to the behavioural cloning loss, we also utilize the inverse dynamics loss as in ~\cite{dasari2020transformers}. By taking consecutive observation frames $o_t, o_{t+1}, \dots, o_{t+k}$ during training, another MLP will predict inverse actions $a_t, a_{t+1}, \dots, a_{t+k-1}$. The inverse dynamics loss has similar form as~\eqref{loss:bc}:

\begin{align}
    P_i = P\left(a_t, \mu_i(\inx_{o_t}^{\text{attn}}, \inx_{o_{t+1}}^{\text{attn}}), s_i\left(\inx_{o_t}^{\text{attn}}, \inx_{o_{t+1}}^{\text{attn}}\right)\right)  \\ 
    \mathcal{L}_{\text{Inv}} = - \log \left(\sum_{i=1}^m \alpha_i(\inx_{o_t}^{\text{attn}}, \inx_{o_{t+1}}^{\text{attn}}) P_i\right) 
\end{align}
Combing with the contrastive loss $\mathcal{L}_{\text{Rep}}$ introduced in Section~\ref{sec:ssl}, we obtain the overall loss for our method: 
\begin{align}
    \mathcal{L} = \lambda_{\text{Rep}} \mathcal{L}_{\text{Rep}} + \lambda_{\text{BC}} \mathcal{L}_{\text{BC}} + \lambda_{\text{Inv}} \mathcal{L}_{\text{Inv}} 
\end{align}

\section{EXPERIMENTS}
\label{sec:experiments}
\subsection{Task Environment and Dataset}
\label{sec:environment}
 
{\bf Simulation environment}. 
We develop 7 distinct task environments using Robosuite v1.1~\cite{zhu2020robosuite} and combining MetaWorld~\cite{yu2019meta} for additional assets. For each task, we additionally design multiple semantically distinct variations. 
In order to investigate cross-morphology imitation, we also integrate two robot arms. The imitation policy is learned and evaluated on a Panda robot arm but takes a Sawyer robot video as demonstration. 

{\bf Data collection}. For every variation of each task environment, we design scripted expert policies and collect 100 demonstration videos of Sawyer robot and another 100 for Panda robot, with differently initialized scene layouts as instances. 
We provide more detailed information on simulation environment and data collection in Appendix. 



\subsection{Experimental Results}

\begin{table*}[t]
\footnotesize
\setlength\tabcolsep{4.0pt}
\centering
\begin{tabular}{ l l r c c c c c} \toprule \toprule 
Task & 
Setup &
DAML~\scriptsize{\cite{yu2018one}} & 
 
\footnotesize{T-OSIL~\scriptsize{\cite{dasari2020transformers}}} &
LSTM & 
MLP & 
MOSAIC \scriptsize{(ours)} \\ 
\cmidrule{1-7}\morecmidrules \cmidrule{1-7}

\multirow{2}{*}{Door} & \footnotesize{single} &	
23.3	$\pm$	5.2	&	57.9 $\pm$	7.1	&	65.8 $\pm$	7.1	&	41.2	$\pm$	8.2	&	{\bf67.1	$\pm$	5.5}	\\ 
  
& \footnotesize{multi} & 
10.8    $\pm$   5.4 & 49.2  $\pm$ 6.0 & 43.8 $\pm$ 9.5  & 58.8 $\pm$ 7.1  & {\bf 68.3 $\pm$ 6.3}  
\\	
\midrule

\multirow{2}{*}{Drawer} & single &	
15.4	$\pm$	5.5	&	57.5	$\pm$	3.9	&	57.5	$\pm$	8.1	&	57.9	$\pm$	3.6	&	{\bf 65.4	$\pm$	3.4} \\ \cmidrule{3-7}
& multi & 
3.3  $\pm$ 1.4 & 53.3 $\pm$ 4.0 & 28.7 $\pm$ 6.0  & 52.5  $\pm$ 6.0 & {\bf 55.8 $\pm$ 3.6} 
\\ 
\midrule 
			
\multirow{2}{*}{Press  Button} & single  &																	62.8	$\pm$	3.9	&	56.4	$\pm$	2.4	&	48.3	$\pm$	6.6	&	40	$\pm$	5.5	&	{\bf 71.7	$\pm$	3.9}	\\ \cmidrule{3-7}
& multi & 
1.7  $\pm$ 0.7 & 63.3 $\pm$ 3.5 & 25.8 $\pm$ 3.0  & 25.0  $\pm$ 3.8 &  {\bf 69.4 $\pm$ 3.4}
\\ \toprule

\multirow{2}{*}{Pick \& Place} & single &																0	$\pm$	0	&	74.4	$\pm$	2.1	&	10.6	$\pm$	1.8	&	12.8	$\pm$	2.3	&	{\bf 88.5	$\pm$	1.1} \\  \cmidrule{3-7}
& multi & 
0.0 $\pm$ 0.0 &  19.5 $\pm$ 0.4 &  2.2 $\pm$ 0.7  &  5.0  $\pm$ 1.4 & {\bf 42.1 $\pm$ 2.3}
\\ \toprule

\multirow{2}{*}{Stack Block} & single &
10.0	$\pm$	1.8	&	13.3	$\pm$	2.6	&	8.6	$\pm$	2.3	&	52.5	$\pm$	4.7	&	{\bf 79.3	$\pm$	1.8}	\\  \cmidrule{3-7}
& multi &
0.0 $\pm$ 0.0 & 34.4 $\pm$ 3.4 & 33.3 $\pm$ 5.5  & 16.7  $\pm$ 3.7 & {\bf 70.6 $\pm$ 2.4} 
\\ \toprule

\multirow{2}{*}{Basketball} & single &
0.4	$\pm$	0.3	&	12.5	$\pm$	1.6	&	5.4	$\pm$	1.2	&	24.2	$\pm$	2.6	&	{\bf 67.5	$\pm$	2.7}	\\  \cmidrule{3-7}
& multi &
0.0 $\pm$ 0.0 &  6.9 $\pm$ 1.3 & 12.1 $\pm$ 2.1  & 10.0  $\pm$ 2.0 & {\bf 49.7 $\pm$ 2.2}
\\ \toprule

\multirow{2}{*}{Nut Assembly } & single &
2.2	$\pm$	1.4	&	6.3	$\pm$	1.9	&	3.9	$\pm$	1.5	&	15.6	$\pm$	2.9	&	{\bf 55.2	$\pm$	2.8}  	\\  \cmidrule{3-7}
& multi &
0.0 $\pm$ 0.0 &  6.3 $\pm$ 1.3 &  4.4 $\pm$ 1.3  & 6.7  $\pm$  1.3 & {\bf 30.7 $\pm$ 2.5} 
\\  

\bottomrule \bottomrule 
\end{tabular}
\vspace{0.15in}
\caption{
Test-time one-shot imitation performance as measured by success rate (\% ) on both single-task and multi-task setup. For each task, 1) each entry of the row named ``single" reports results of a single-task model that was trained and tested on the same task; 2) the row named ``multi" reports results from one multi-task model that was trained on all 7 tasks in the benchmark and tested on each task separately.
}
\label{tab:both-single-multi}
 
\end{table*}
 
We conduct experiments with the dataset described in Section~\ref{sec:environment} to answer the following questions:
\begin{itemize}
    \item How does our method compare with prior baselines under the original single-task one-shot imitation learning setup.
    \item How well does our method perform across multiple tasks, after trained on the same set of tasks.
    \item How well does our trained multi-task model perform given a completely new task: can it 1) directly perform one-shot imitation at test time; 2) be fine-tuned to quickly adapt to the new task requiring fewer amount of data.
    \item Which component(s) in our contrastive module are key to its effectiveness at representation learning. The ablation experiment results are provided in Appendix due to space limitation.
\end{itemize}
We report and compare performance to the following baseline methods:
\begin{itemize}
    \item \textbf{DAML}~\cite{yu2018one}: We train a policy model with MAML~\cite{finn2017model} loss and behaviour cloning loss. We replace the model architecture used in the original paper with a wider and deeper network of comparable parameter counts as ours, and use the same action distribution parametrization for policy learning.
    \item \textbf{T-OSIL}~\cite{dasari2020transformers}:  
     Model architecture proposed by \cite{dasari2020transformers} which also uses non-local block~\cite{wang2018non} in self-attention module and is trained with end-effector point prediction as auxiliary loss. Our method utilizes a more computationally efficient attention operation, see Figure \ref{fig:arch-ours} for an illustrated comparison. 
    \item \textbf{LSTM}: We replace the self-attention module in our model architecture with linear projectors followed by an LSTM~\cite{hochreiter1997long, sutskever2014sequence} architecture of a comparable parameter count. The rest of the policy model architecture is kept the same as in our main method.
    \item \textbf{MLP}: We replace the self-attention module in our model architecture with a simple MLP layer to process stacked visual features from the demonstration into ``task context vectors", which is then concatenated with observation features and used for action predictions.   
\end{itemize}
For all experiments (including baselines), we keep the first three convolutional residual blocks in ResNet-18 as the feature extractor, and apply the same data augmentation strategy to prevent over-fitting and improve model robustness. For evaluation, we take 3 different converged model checkpoints, and for each variation of each task, we gather each policy's rollout performance across 10 episodes with different random seeds. For both the single and multi-task models, we report the mean and standard deviation of success rates in each task separately. Details on network architecture and hyper-parameters can be found in the Appendix.

\textbf{Single-task One-shot Imitation}

We first evaluate performance on the single-task setup as done in prior methods. Specifically, the model is trained with demonstrations from multiple variations of one task and then tested on unseen instances of the same task (e.g. different initial poses). We report the success rate of all methods on each task in the rows named ``single" of Table~\ref{tab:both-single-multi}, with a clear out-performance of our method over baselines on every task. 

We remark that DAML \cite{yu2018one} also experimented with a pick-and-place task in their original paper, but collected ``hundreds of" objects for training and 12 held-out objects for testing. Without access to further details, we hypothesize that this visual diversity in training dataset was crucial for its success at picking correct test-time objects, which explains why the same method reports massive under-performance on the 4-object Pick \& Place task, where \cite{dasari2020transformers} also reported a very low success rate from DAML ($6.9\%$) using a differently-configured simulation of the same task.

\textbf{Multi-task One-shot Imitation}
 
We next consider the multi-task setup by mixing data from all 7 tasks into one training dataset. After training, we report the success rate of one model on each task in the rows named ``multi" of Table~\ref{tab:both-single-multi}. 
The baselines' performances drop significantly compared with their results from single-task, whereas ours continues to work well across many tasks \footnote{For this setup, we increase the number of attention layers in the model from 2 to 3, and also adjust each baseline accordingly for fair comparisons. However, the performance of DAML~\cite{yu2019meta} is still not comparable with others using the updated architecture and after tuning 
hyper-parameters.}.


\textbf{Novel Task Generalization}
\begin{figure} [t]  
\includegraphics[width=.5\textwidth]{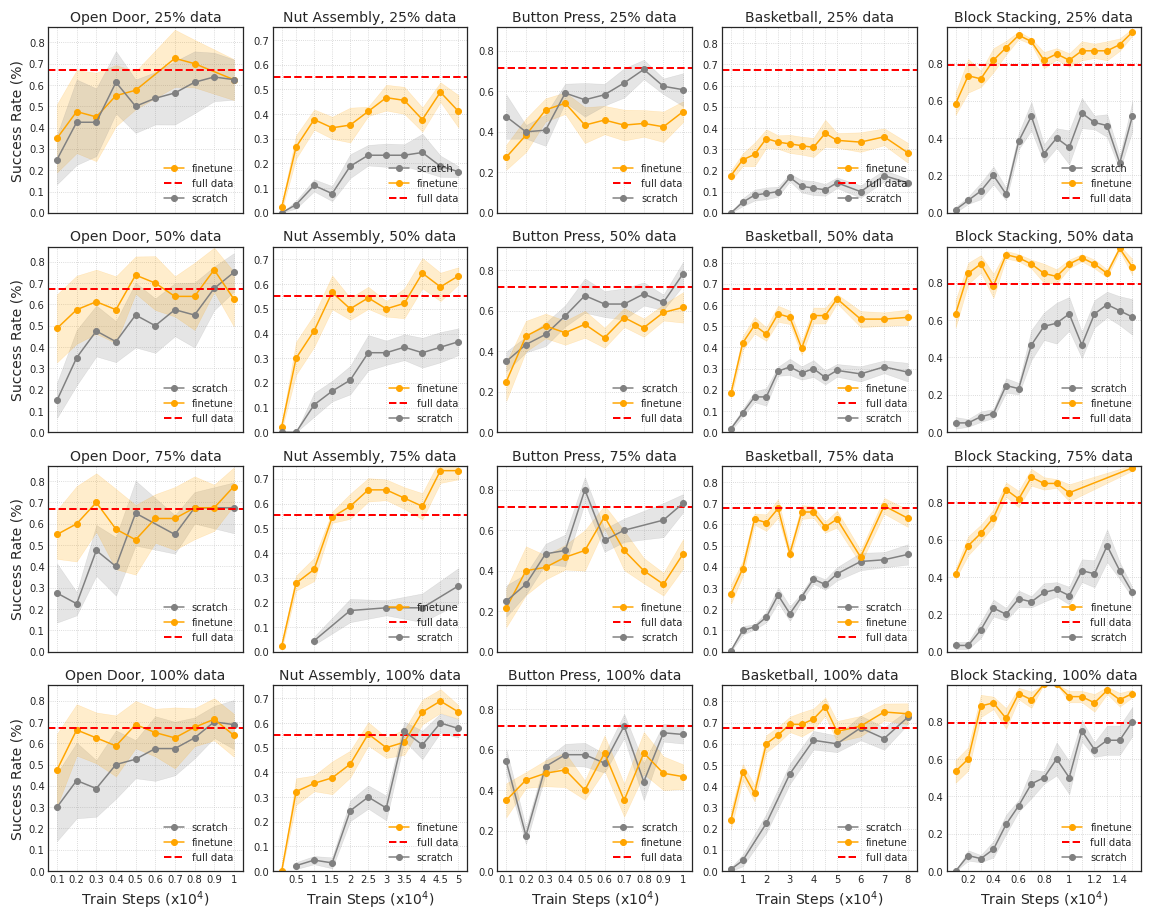}
\caption{\small We compare fine-tuning multi-task models on their corresponding held-out novel task versus training a single-task model from scratch. We intermittently save model checkpoints throughout each training run and plot results for evaluating one-shot imitation performance. To additionally compare the data efficiency between fine-tuning and train-from-scratch, we experiment with restricted amount of demonstration data used for training, i.e. 25\%, 50\% and 75\% of the data that was used for single-task and multi-task experiments and reported results in Table \ref{tab:both-single-multi}
}
\label{fig:finetune}
\end{figure}

\begin{table*}[htbp!]
\footnotesize
\centering
\begin{tabular}{l|r|r|r|r|r}\toprule

Training Setup &
Novel-task &
No Training  &
Single-task & 
Multi-task & 
Fine-tune  

 \\ \midrule

Door         & 5.0  $\pm$ 3.1  & 2.5 $\pm$ 0.9  &	67.1  $\pm$	5.5	&  68.3	 $\pm$	6.3	&	67.5  $\pm$	5.6	    \\

Drawer       & 15.0 $\pm$ 6.6  & 1.2 $\pm$ 1.2  & 65.4  $\pm$ 3.4	&  55.8  $\pm$  3.6	&   52.5  $\pm$   4.5   \\																									
Press Button & 5.0	$\pm$ 3.1  & 0              & 73.9  $\pm$ 3.9	&  69.4	 $\pm$	3.4	&	51.1    $\pm$   4.8	 \\
																					
Stack Block	 & 	0	           & 0              & 79.3  $\pm$ 1.8	&  70.6	 $\pm$	2.4	&   98.3	$\pm$   0.9  \\
																									
Basketball	 &  0	           & 0              &  	67.5  $\pm$	2.7 &  49.7  $\pm$	2.2	&	72.8	$\pm$   2.7	 \\
																									
Nut Assembly &	0	           & 0              &  	55.2  $\pm$	2.8	&  30.7	 $\pm$	2.5	&   73.3	$\pm$   2.3	  \\

\bottomrule 
\end{tabular}
\vspace{0.1in}
\caption{
\small{
Evaluation success rate (\% ) of one-shot imitation with MOSAIC under different train/test settings. \textbf{Novel-task}: directly evaluate on the task at each row with a model that was only trained on all of the remaining tasks. \textbf{No Training}: evaluate a randomly initialized policy model without any training. \textbf{Single-task}: train and test on the same task in each row. \textbf{Multi-task}: train a model on all 7 tasks, then evaluate and report the final performance on each task separately. 
}
}
\label{tab:zero-shot}
\end{table*}  
To show a higher level of generalization ability, we test a one-shot imitation agent with tasks that are sufficiently different from what it already trained on. We hence set up a series of experiments that, each picks 1 out of the 7 tasks in our benchmark suite as the held-out task, and trains a multi-task model on the remaining 6 tasks until convergence. 

We first directly evaluate each model on its corresponding held-out task. Results are reported in column ``Novel-task'' of Table \ref{tab:zero-shot}, where each row corresponds to the experiment where the current task was excluded from training and only used for one-shot evaluation. For comparison purposes, we also include: ``No Training" column, which evaluates a randomly initialized policy network without any training , ``Single-task'' column, where each row reports performance of a model trained and tested on the same task, and ``Multi-task'' column, where we take one model trained on all 7 tasks and report its performance on each task separately.
 
As shown in Table \ref{tab:zero-shot}, directly evaluating a multi-task model largely fails to complete an unseen novel task, and performs significantly worse than when this novel task was included during single- or multi-task training.  This failure of direct one-shot imitation on a novel task suggests exciting room for future research.
Nevertheless, the gap between ``Novel-task" and ``No Training" results \footnote{
We remark that, the non-zero success rates in ``No Training" column are due to the nature of task design in simulation, where a randomly initialized policy model would sometimes generate an action that accidentally leads to an episode being counted ``successful", such as hitting an opened drawer to shut it close or stumbling on a closed door and pushing it open.
} suggests a multi-task model still learns certain non-random behaviors, which could be generalized directly to a completely novel task. Note that, the improvement over ``No Training" is limited to the first three tasks (Door, Drawer and Press Button), which require simpler motions and task reasoning.

We are hence encouraged to continue from this setup, but further fine-tune each 6-task model on its corresponding held-out task. We use the exact same dataset for training the models in ``Single-task" column, and report the final fine-tuned results in ``Fine-tune" column. To investigate the data efficiency of this setup, we additionally experiment with using $25\%$, $50\%$, $75\%$ the amount of the original single-task training data for fine-tuning, and compare with training from scratch on that single task using the same dataset size. 

We plot evaluation results of intermittently-saved checkpoints during each model's training in Figure~\ref{fig:finetune}. We observe that a multi-task pre-trained model is able to adapt quickly to a completely new task, even when limited data is available (i.e., $25\%$), and the final convergence performance is sometimes higher than training single-task from-scratch for some challenging tasks (e.g. Nut Assembly, Block stacking and Basketball). This indicates that these models are indeed able to accumulate generalizable knowledge (such as flexible visual feature extractors) from pre-training on many tasks, which then put them at advantage of learning a novel task very efficiently.


\section{CONCLUSION}
In this work, we build on prior progress in one-shot imitation learning and propose a more challenging multi-task setup. Instead of training and testing on different variations/instances of a single task, we call for training with multiple distinct tasks and testing on novel tasks that are never seen during training. We believe this setup is crucial towards building more capable and generalizable agents, and holds great potential for novel research. 

To support this formulation, we introduce a one-shot imitation benchmark for robotic manipulation, which consists of 61 variations across 7 different tasks. We propose our method MOSAIC, which combines a self-attention model architecture and a temporal contrastive objective, and out-performs previous state-of-the-art methods by a large margin. When evaluated on a completely new task, we see a promising potential in fine-tuning our multi-task model to learn efficiently, but remark the great room for improvement at even better and faster one-shot learning at test time.

\section{ACKNOWLEDGEMENT}
This work was supported by the Bakar Family Foundation, Google LLC, Intel, the Berkeley Center for Human Compatible AI (CHAI), and National Science Foundation grant NRI no. 2024675. The authors would also like to thank a number of lab colleagues for their help throughout the project: Stephen James and Thanard Kurutach for valuable project feedbacks; Aditya Grover, Kevin Lu and Igor Mordatch for insightful discussions; Colin(Qiyang) Li, Hao Liu for suggestions on the paper writing. 

\bibliographystyle{IEEEtranS}
\bibliography{example}

\newpage
\section{APPENDIX}
\subsection{Overview}
In this section, we first provide detailed descriptions of our task environment design, then introduce more details about our method implementation and the experiment setup, and lastly we provide more ablation experiment results and analysis. 

\subsection{Task Environment Description}
We use Robosuite~\cite{zhu2020robosuite} as our base framework, and design 7 robot manipulation environments (tasks) with intra-task variations. For example, consider Nut Assembly task in the bottom-right of Figure~\ref{fig:fig3}: it has 9 variations in total, resulting from picking up any one of the 3 nuts on the table, then assembling it to one of the 3 pegs.  Since there are multiple varied instances for a specific task, the agent should finish the correct task without mis-identification based on demonstration. Below we provide details about each of the individual task design.

\begin{SCfigure}[5][htbp!]
\vspace{-.2cm}
\includegraphics[width=0.16\textwidth]{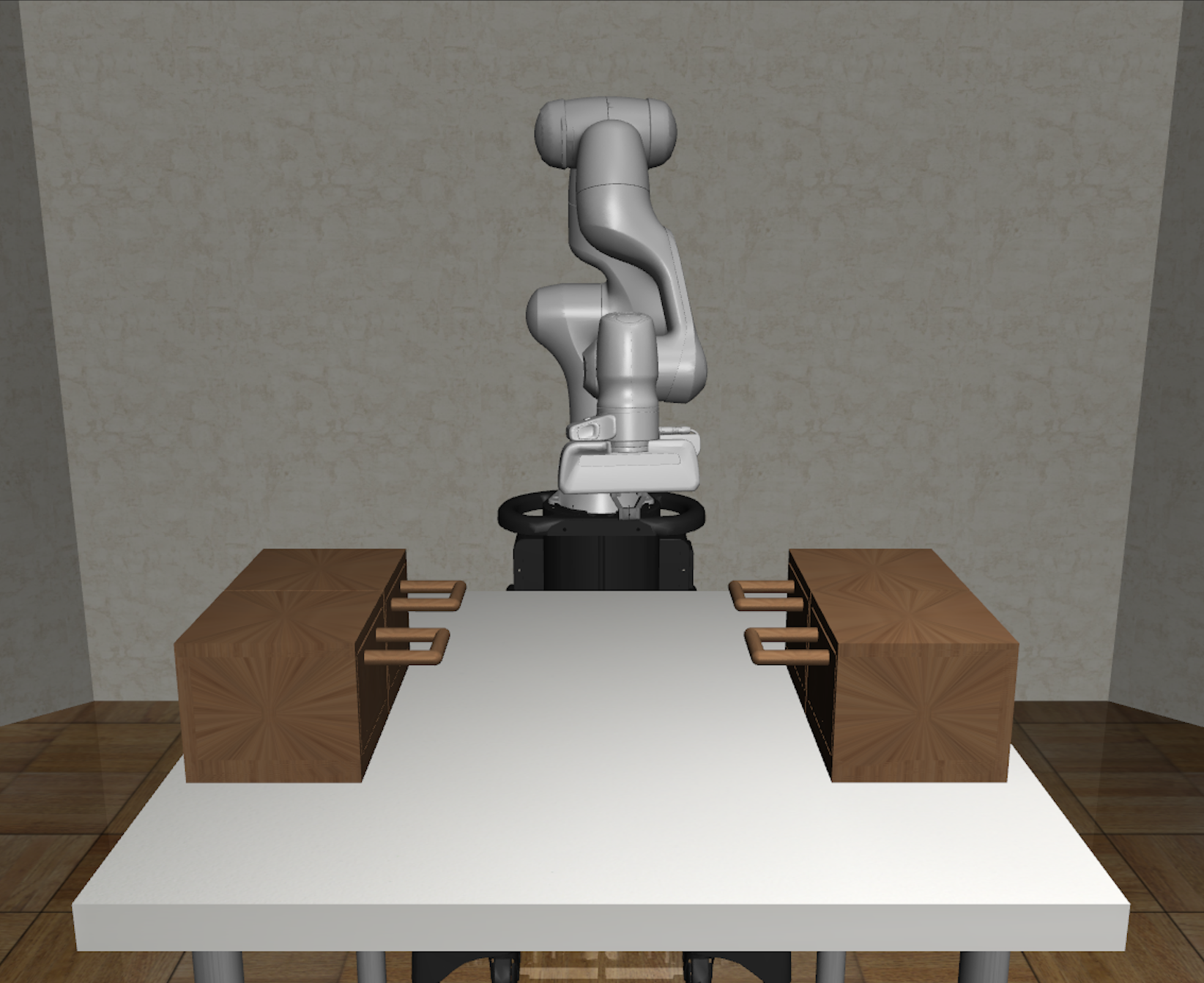} 
\caption{Drawer task with 8 sub-tasks in total. Given a demonstration, the agent should be able to infer which drawer to open or close. Note that all the drawers will be initialized as opened if the task is to close drawer.}
\label{fig:drawer}
\end{SCfigure}
   
\begin{SCfigure}[5][htbp!]
\vspace{-.2cm}
\includegraphics[width=0.16\textwidth]{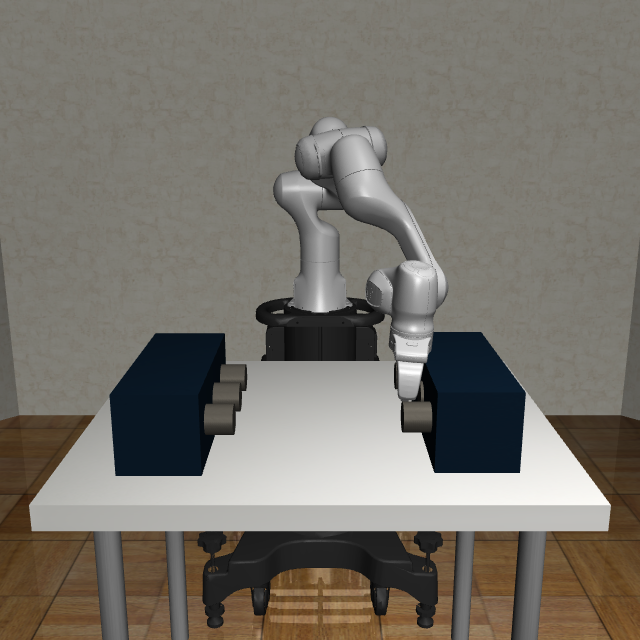}
\caption{Press button with 6 sub-tasks in total. Given a demonstration, the agent should be able to infer which button to press.}
\label{fig:button}
\end{SCfigure}

\begin{SCfigure}[5][htbp!]
\vspace{-.2cm}
\includegraphics[width=0.16\textwidth]{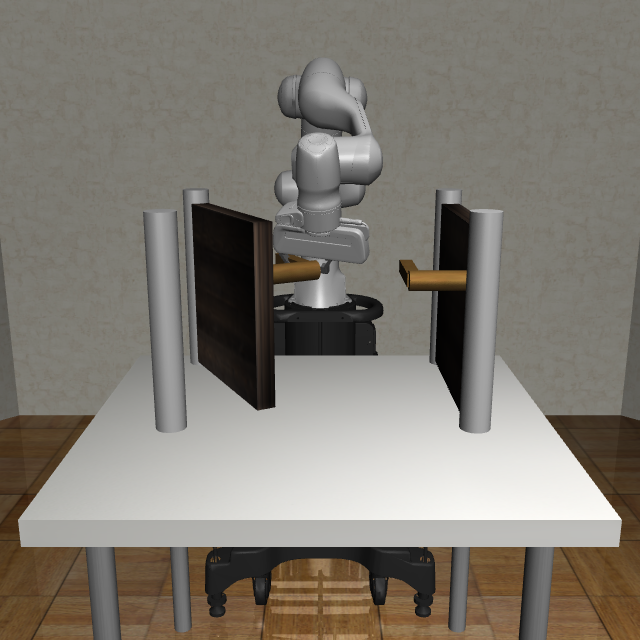}
\caption{Open door with in total 4 sub-tasks. Given a demonstration, the agent should be able to infer which door to open, and whether in clockwise or counterclockwise direction.}
\label{fig:button}
\end{SCfigure}

\begin{SCfigure}[5][htbp!]
\vspace{-.2cm}
\includegraphics[width=0.16\textwidth]{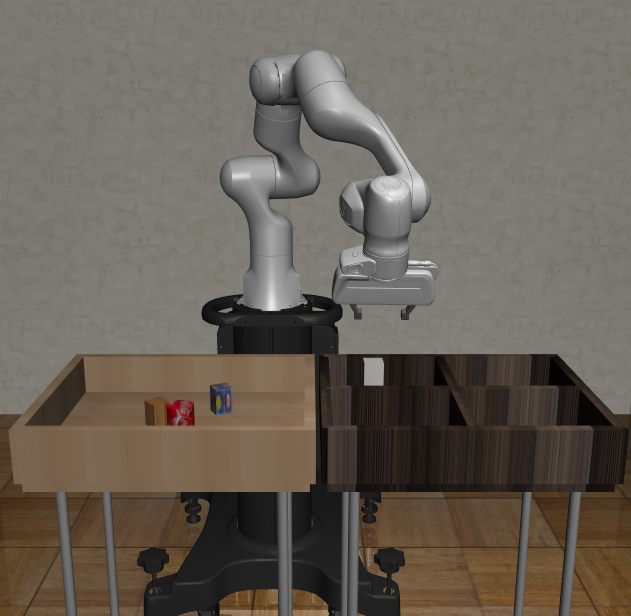}
\caption{Pick \& place environment with in total 16 tasks. Given a demonstration, the agent should be able to infer which object to pick and which bin to place the object in.}
\label{fig:pickplace}
\end{SCfigure}

\begin{SCfigure}[5][htbp!]
\vspace{-.2cm}
\includegraphics[width=0.16\textwidth]{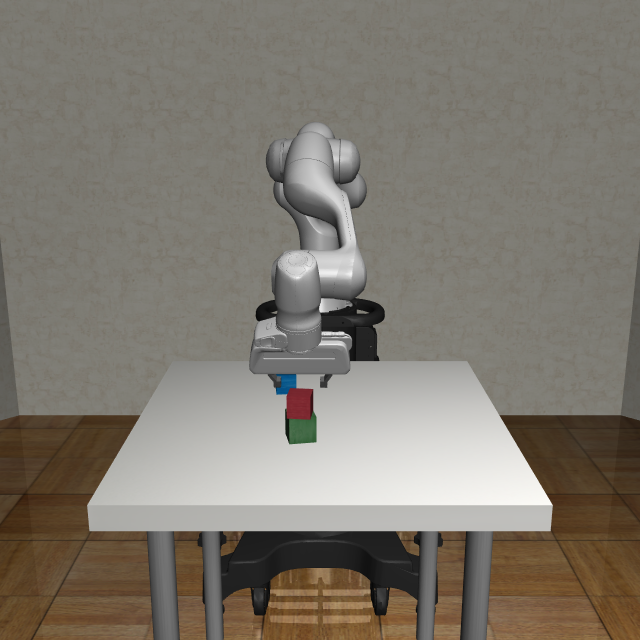}
\caption{Block stacking environment with in total 6 tasks. Given a demonstration, the agent should be able to infer which block to pick and where to stack.}
\label{fig:stack}
\end{SCfigure}

\begin{SCfigure}[5][htbp!]
\vspace{-.2cm}
\includegraphics[width=0.16\textwidth]{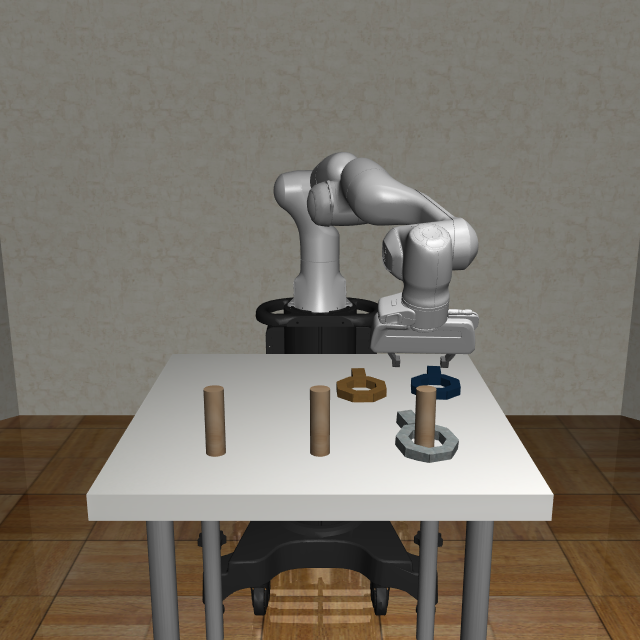}
\caption{Nut assembly environment with in total 9 tasks. Given a demonstration, the agent should be able to infer which nut to pick and which peg to assemble.}
\label{fig:peg}
\end{SCfigure}

\begin{SCfigure}[5][htbp!]
\vspace{-.2cm}
\includegraphics[width=0.16\textwidth]{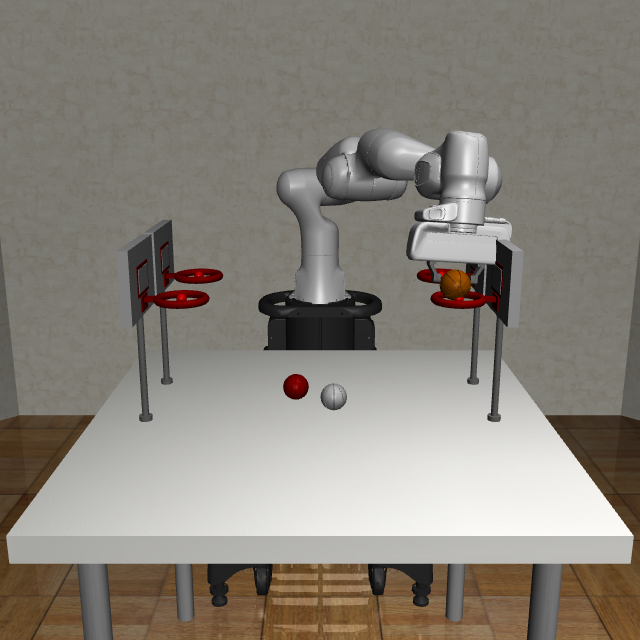}
\caption{Basketball environment with in total 12 tasks. Given a demonstration, the agent should be able to infer which ball to pick and which hoop to throw it in.}
\label{fig:bask}
\end{SCfigure}

\begin{figure*}[htbp!]
\center
\includegraphics[width=0.8\textwidth,]{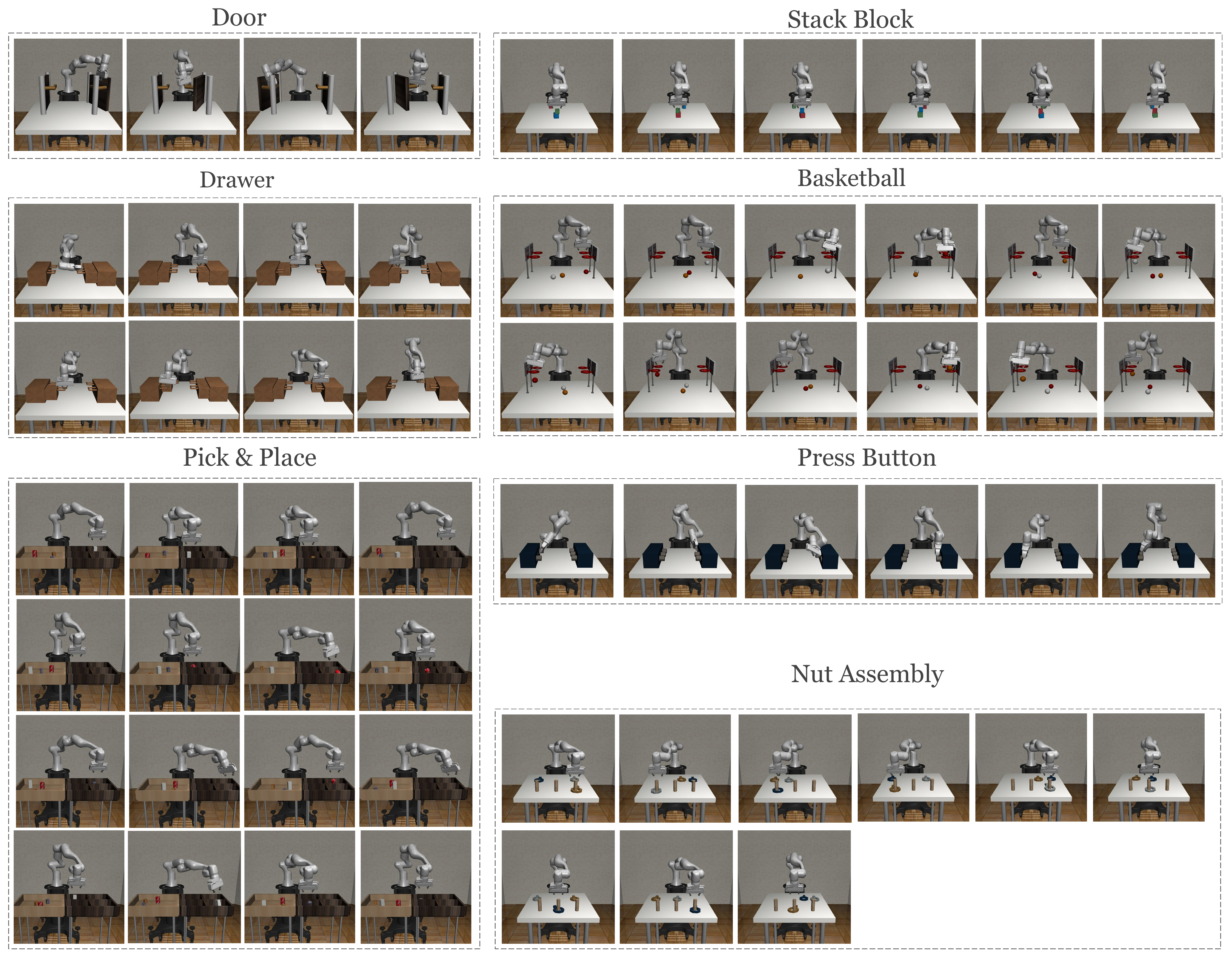}
\caption{
We visualize the entirety of our one-shot robot imitation benchmark of 7 tasks and a total of 61 semantically distinct variations. The number of variations is different across tasks, with a range from 4 (for Door task) to 16 (for Pick $\&$ Place task). For each variation, we also add randomization to create more varied instances, such as different initial object positions. A multi-task one-shot imitation policy is trained on a demonstration dataset that contains multiple tasks and all their variations.
} 
\label{fig:fig3}
\end{figure*}

\subsection{Implementation Details}
\label{sec:appendexB}
\begin{figure}[h]
\vspace{-.1cm}
\centering
\includegraphics[width=0.4\textwidth]{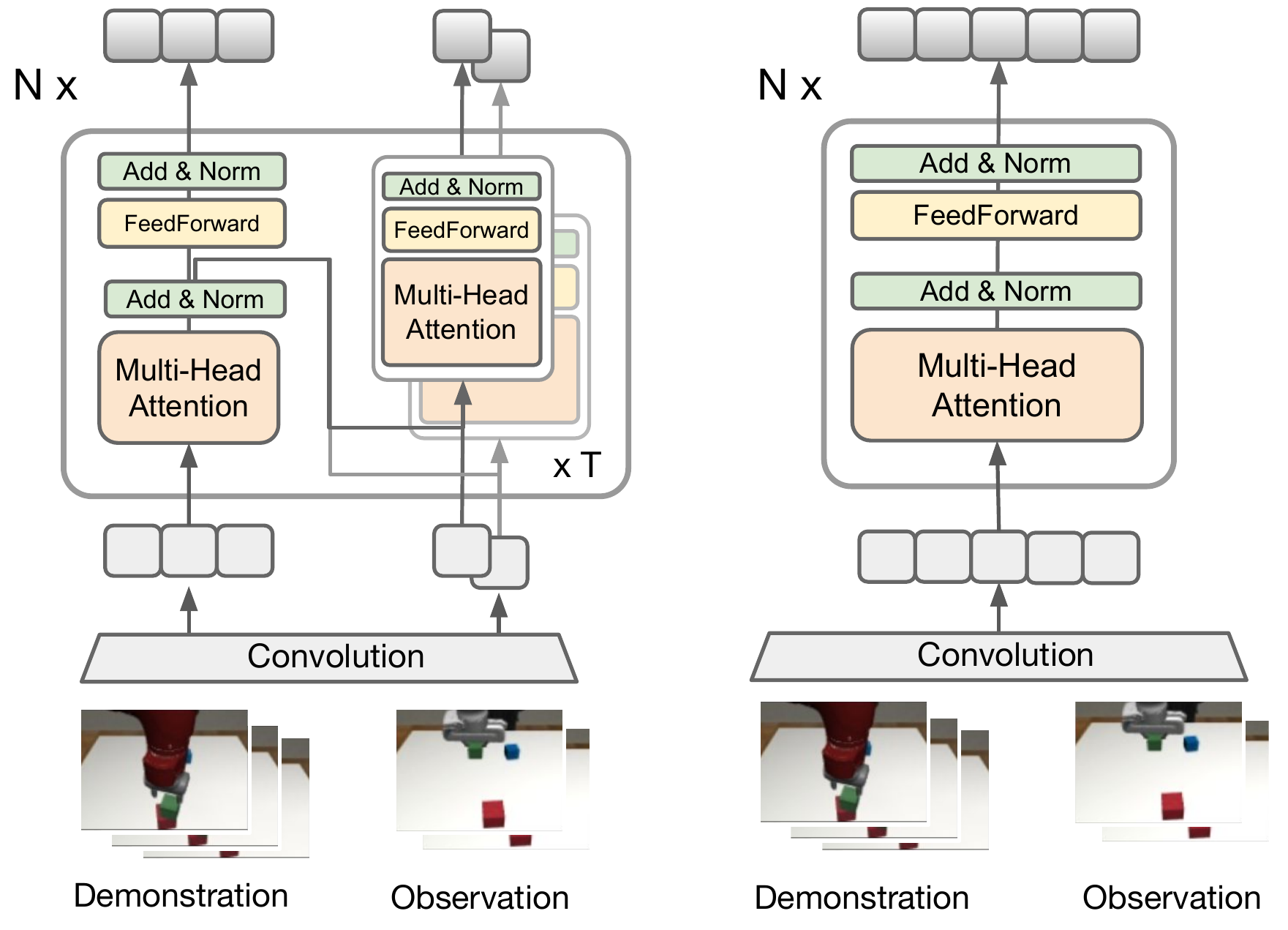}
\caption{Illustration of self-attention blocks used in our model (left) architecture and T-OSIL~\cite{dasari2020transformers} (right). By removing the attention calculation between frames within the observation input sequence, we significantly reduces the computational overhead and makes our model architecture able to scale better to longer length of observations. At test time, only one frame is given to model inference per each time-step, which make for faster evaluation and gaining the same or better performance than stacking multiple frames instead.}
\label{fig:arch-ours}
\end{figure}

\textbf{Network Architecture}
We provide an illustration of our self-attention block and one of our baseline~\cite{dasari2020transformers} as in Figure~\ref{fig:arch-ours}. For other baselines in our main paper, we replace this attention module with LSTM or MLP accordingly.

\textbf{Data Augmentation}
We add data augmentation for all the baselines to prevent overfitting. In our implementation, we use 4 types of data augmentation provided in the {\em torchvision} package: random translate, random crop, color jitter, Gaussian blur.

\subsection{Experiment Details}

\label{sec:appendexC}
{\bf Hyper-parameter Settings}
We provide more details about the hyper-parameters and other settings of model training and evaluation in Table~\ref{tab:hyperparam}.
\begin{table}[ht]
\vskip 0.15in
\begin{center}
\begin{small}
\begin{tabular}{ll}
\toprule
\textbf{Hyperparameter} & \textbf{Value}  \\
\midrule
Input image size    & $(100, 180, 3)$  \\ 
\# Demonstation frames & 4 \\
\# Observation frames (train) & 7 \\
\# Observation frames (eval) & 1 \\
\# Evaluation episode per task  & $10$  \\ 
Optimizer    & Adam  \\ 
$(\beta_1,\beta_2)$  & $(.9,.999)$  \\
Learning rate  & $5e^{-4}$ \\
Batch size    & $30$   \\ 
Non-linearity & ReLU \\
Contrastive latent dimension & $512$ \\
Self-attention temperature & $16$ \\
\# Action layers & $2$ \\ 
\# Attention layers (single-task) & $2$ \\ 
\# Attention layers (multi-task) & $3$ \\ 
Action head latent dimension & 256 \\
Action output dimension & 256 \\

\bottomrule
\end{tabular}
\end{small}
\caption{Hyperparameters used for model training and evaluation.}
\label{tab:hyperparam}
\end{center}
\vskip 0.1in
\end{table}

{\bf Batch Construction}
One batch consists of demonstrations sampled from each variant in every training task, which are mixed evenly. Since each task contains a different number of variations, to ensure they have comparable learning progress, the loss is first averaged across variants within each task, and then across different tasks.

{\bf Computation Requiremenmts}
Our single-task model can be trained within one GPU day of NVIDIA TITAN Xp, whereas the multi-task model takes about 4 GPU days to train.

\subsection{Ablation Experiments}
\begin{table*}[t]
\centering
\setlength\tabcolsep{4.0pt}
\footnotesize
\small
\begin{tabular}{l|ccccccccc} \toprule
\begin{tabular}[c]{@{}c@{}} Task  \end{tabular}
& \begin{tabular}[c]{@{}c@{}} Door \end{tabular}
& \begin{tabular}[c]{@{}c@{}} Drawer \end{tabular}
& \begin{tabular}[c]{@{}c@{}} Press \\ Button \end{tabular} 
& \begin{tabular}[c]{@{}c@{}} Pick \& \\ Place \end{tabular}  
& \begin{tabular}[c]{@{}c@{}} Stack \\ Block \end{tabular} 
& \begin{tabular}[c]{@{}c@{}} Basketball \end{tabular} 
& \begin{tabular}[c]{@{}c@{}} Nut \\ Assembly \end{tabular}  \\ \midrule
 
\begin{tabular}[c]{@{}c@{}} Single-task \\ w/ Contra. \end{tabular} 
& 67.1 $\pm$ 5.5  & 65.4 $\pm$ 3.4     & 71.1 $\pm$ 3.9     & 88.8 $ \pm$ 1.1      & 79.3 $\pm$ 1.8       & 67.5 $\pm$ 2.7   & 55.2 $\pm$ 2.8 \\  \midrule

\begin{tabular}[c]{@{}c@{}} Single-task \\ w/o Contra. \end{tabular}
& 62.5	$\pm$ 8.2   & 60.8 $\pm$ 2.6  &  73.9 $\pm$ 3.9 & 69.5 $\pm$	1.8    & 36.7 $\pm$	3.6  & 19.7	$\pm$ 2.2       & 15.6	$\pm$	2.5 \\  \midrule

\begin{tabular}[c]{@{}c@{}} Multi-task \\ w/ Contra. \end{tabular}
& 68.3 $\pm$ 6.3    & 55.8 $\pm$ 3.6  & 69.4 $\pm$ 3.4  & 42.1 $\pm$ 2.3   & 70.6 $\pm$ 2.4  & 49.7 $\pm$ 2.2 & 30.7 $\pm$ 2.5 \\  \midrule 

\begin{tabular}[c]{@{}c@{}} Multi-task \\ w/o Contra. \end{tabular}
& 66.7 $\pm$ 7.3    & 67.5 $\pm$ 4.0  & 73.1 $\pm$ 3.4  & 11.9 $\pm$ 1.6    & 39.4 $\pm$ 3.4    & 9.2  $\pm$ 1.7      & 20.4 $\pm$ 1.9 \\  \midrule 
 
\begin{tabular}[c]{@{}c@{}} Multi-task \\ BYOL \end{tabular}
& 67.1 $\pm$ 6.4    & 67.9 $\pm$ 4.0  & 69.4 $\pm$ 3.1    & 12.9 $\pm$ 1.7    & 40.0 $\pm$ 2.4    & 10.0 $\pm$ 1.9      & 17.8 $\pm$ 1.8 \\ 

\bottomrule
\end{tabular}
\vspace{0.1 in}
\caption{\small Ablation studies: the effectiveness of contrastive learning with negative samples, as measure by success rates on single- and multi-task setups. We remark that 1) contrastive loss improves one-shot imitation performance on both setups, and the gain is most significant for multi-task models and on the last three tasks (Stack Block, Baseketball, and Nut Assembly); 2) contrastive learning without negative samples (as reported in the row ``Multi-task BYOL") results in similar performance as not using contrastive module at all, which is reported in row ``Multi-task w/o Contra.".}
\label{tab:ablation-byol}
\end{table*}

\begin{table*}[t]
\small
\centering
\begin{tabular}{l|ccccc}\toprule
\begin{tabular}[c]{@{}c@{}} Task Domain  \end{tabular}
& \begin{tabular}[c]{@{}c@{}} No-Temp \end{tabular}
& \begin{tabular}[c]{@{}c@{}} Fix-Temp   \end{tabular}
& \begin{tabular}[c]{@{}c@{}} Pre-Attn    \end{tabular}
& \begin{tabular}[c]{@{}c@{}} Post-Attn    \end{tabular}
& \begin{tabular}[c]{@{}c@{}} \text{\scriptsize Rand-Temp +} \\ \text{\scriptsize Both-Attn} \end{tabular} \\ \midrule 
Pick \& Place  & 12.3 $\pm$ 1.4  & 80.2 $\pm$ 1.9 & 23.5 $\pm$ 2.1    & 51.5 $\pm$ 4.1    & \bf{88.5 $\pm$	1.1}    \\   

Basketball & 53.8 $\pm$ 4.8  & 52.5 $\pm$ 2.8 & 24.7 $\pm$ 2.1    & 27.2 $\pm$ 2.5    & {\bf 67.5	$\pm$	2.7}   \\   

Nut Assembly   & 12.2 $\pm$ 2.0  & 12.2 $\pm$ 1.6 & 17.8 $\pm$ 1.8    & 22.6 $\pm$ 2.9    & {\bf 55.2	$\pm$	2.8}   \\   \midrule 
\end{tabular} 
\vspace{0.1 in}
\caption{\small Ablation studies: (1) whether and how to enforce temporal consistency (Column 1, 2, 5); (2) Where to apply contrastive loss throughout the model (Column 3-5).} 
\label{tab:ablation}
\end{table*}
\label{sec:ablation}

We conduct a set of ablation studies focused on our contrastive representation learning approach, to provide insights on the design of the loss objective, how it incorporates with the policy learning, and its impact over the one-shot imitation performance. 

{\bf The effect of contrastive learning objective} 

We first ablate by removing contrastive loss from the model's training update: we train a model with solely behavior cloning loss, while other training configurations are kept consistent with the multi-task experiments reported in Table \ref{tab:both-single-multi}. Results are reported in row ``Single-task No Contra."  and ``Multi-task No Contra." of Table~\ref{tab:ablation-byol}. Noticeably, our multi-task performance is significantly decreased without the contrastive learning objective. Comparing results from single-task and multi-task setups, we observe a clear challenge of task reasoning and robust representation learning, which the contrastive objective is able to address, but not fully closing the gap between a multi-task model and its single-task counterpart on every task. 

Adding the contrastive loss also shows a more significant gain on the right-most 3 tasks of Table~\ref{tab:ablation-byol}, which vary only the colors of otherwise similarly-shaped objects. As compared to Pick \& Place where the 4 objects are differently sized, shaped, and colored, the blocks/basketballs/nuts in these variations tend to be more easily confused, which can be addressed the contrastive loss that explicitly forces the representations to be distinguishable among different sub-tasks. 

To investigate the effect of contrasting against negative samples in the loss objective design, we implement BYOL~\cite{grill2020bootstrap}, in which a data sample is only drawn together with its augmented counterpart. Its multi-task performance is shown in Table~\ref{tab:ablation-byol}, which show no improvement over not using contrastive loss at all (as reported in row ``Multi-task No Contra."). We remark that in multi-task setups, negative samples help learn stronger representations that are better at distinguishing among images from different tasks/variations.

{\bf Where to apply contrastive loss}. 
As discussed in Section~\ref{sec:ssl}, 
we can apply the contrastive operation in features from any intermediate layers of the model. To understand the effects of this algorithmic choice, we consider the following variants of our method:
(1) {\em Pre-Attn}: applying contrastive loss to only features prior to the self-attention layers;
(2) {\em Post-Attn}: applying contrastive loss only after the self-attention layers; 
(3) {\em Both-Attn (Ours)}: calculating both losses in (1) and (2), which we use for single and multi-task experiment results in above sections.
Table~\ref{tab:ablation} shows the performance of each ablation variant in three single tasks, among which we find that {\em Both-Attn} (using two contrastive losses) achieves the best performance.

{\bf The effect of temporal contrast}. 

Our contrastive strategy is doing a {\em temporal} contrast by {\em randomly} selecting frames from nearby time-steps as positive. To fully understand its effectiveness,
we compare it with two variants:
(1) {\em No-Temp}: applying contrastive loss to two different data augmentations from same frame, similar to CURL~\cite{laskin2020curl};
(2) {\em Fix-Temp}: applying temporal contrast but always with a fixed-step future frame, similar to ATC~\cite{dwibedi2019temporal}. Ours (denoted {\em Rand-Temp}) achieves the best performance as shown in Table~\ref{tab:ablation}, as suggested by comparing the first, second and last column in Table~\ref{tab:ablation}.

\subsection{Further Discussion on Related Work}
In this section, we provide a more detailed discussion of the methods and experiment task settings used by related prior work in one-shot imitation learning (OSIL).   

However, prior OSIL work has been limited to a {\em single-task} setup and mainly tests a model on a different task variation (e.g stacking an unseen block combination) or a different instance (e.g. different object pose) of the previously-seen variations. Experiments in \cite{duan2017one} train an agent to stack various (unseen) block combinations at test time, but use low-dimensional state-based inputs. For visual inputs, \cite{finn2017one} and \cite{james2018taskembedded} experimented with 3 separate settings: simulated planer reaching (with different target object colors), simulated planer pushing (with varying target object locations), and real-robot, object-in-hand placing (onto different target containers); \cite{yu2018one} set up a two-stage pick-then-place task with varying target objects and target containers; \cite{dasari2020transformers} uses a simulated Pick \& Place task  with 4 objects to pick and 4 target bins to place (hence 16 variations in total). The AI2-THOR \cite{kolve2019ai2thor}  environment used in \cite{Huang2019ContinuousRO} requires collecting varying objects and dropping off at their designated receptacles, where actions are purely semantic concepts such as ``dropoff" or ``search". In contrast, in this work we consider a harder, multi-task setup, where agent needs to perform well across more diverse and distinct tasks, and generalize not only to new instances of all the seen variations, but also to completely novel tasks. 

\end{document}